# Can Large Language Models Bridge the Gap in Environmental Knowledge?

Linda Smail[1], David Santandreu Calonge[2], Firuz Kamalov[3], Nur H. Orak[4]

[1]College of Interdisciplinary Studies, Zayed University, Dubai, United Arab Emirates, Linda.Smail@zu.ac.ae, https://orcid.org/0000-0001-9388-1334

[2]Department of Academic Development, Mohamed bin Zayed University of Artificial Intelligence, Abu Dhabi, United Arab Emirates, david.santandreu@mbzuai.ac.ae, https://orcid.org/0000-0003-0101-8758

[3]School of Engineering, Applied Science and Technology, Canadian University Dubai, United Arab Emirates, firuz@cud.ac.ae, https://orcid.org/0000-0003-3946-0920

[4]Department of Environmental Engineering, Marmara University, Istanbul, Türkiye, nur.orak@marmara.edu.tr, https://orcid.org/0000-0002-3830-9260

*Corresponding author: Dr. Linda Smail, Computational Systems Program, College of Interdisciplinary Studies, Zayed University, United Arab Emirates, Linda.Smail@zu.ac.ae, https://orcid.org/0000-0001-9388-1334

**Abstract:**

This research investigates the potential of Artificial Intelligence (AI) models to bridge the knowledge gap in environmental education among university students. By focusing on prominent large language models (LLMs) such as GPT-3.5, GPT-4, GPT-4o, Gemini, Claude Sonnet, and Llama 2, the study assesses their effectiveness in conveying environmental concepts and, consequently, facilitating environmental education. The investigation employs a standardized tool, the Environmental Knowledge Test (EKT-19), supplemented by targeted questions, to evaluate the environmental knowledge of university students in comparison to the responses generated by the AI models. The results of this study suggest that while AI models possess a vast, readily accessible, and valid knowledge base with the potential to empower both students and academic staff, a human discipline specialist in environmental sciences may still be necessary to validate the accuracy of the information provided.

1. **Introduction**

Extreme weather events, increasing global temperatures, rising sea-levels, and changes to ecosystems and biodiversity are all consequences of climate change, which is mostly caused by anthropogenic greenhouse gas emissions (Masson-Delmotte et al., 2018). Meanwhile, the loss of biodiversity due to habitat degradation, pollution, overexploitation, and invasive species threatens the resilience of society's ecosystems (Nature, 2021). These consequences pose questions regarding food security, public health, and socioeconomic stability. Thus, effective access to accurate environmental knowledge is crucial for developing sustainable solutions and informed environmental policies.

Understanding the value of environmental knowledge requires awareness of the various environmental issues facing humanity today. In this article, environmental knowledge is described as the understanding of environmental issues and ideas such as natural systems, climate change, pollution, biodiversity, ecosystems, and how human actions affect them. Unfortunately, many people are ignorant of these issues and their consequences. Despite the critical role of environmental knowledge, there are significant gaps in access and dissemination. Traditional channels for sharing environmental knowledge, including academic papers, expert consultations, and reports, often lack the reach, simplicity, and accessibility necessary for broad public engagement. Moreover, the interdisciplinary complexity of environmental issues further complicates their practical application. Consequently, limited environmental knowledge among stakeholders can hinder informed decision-making and reduce the effectiveness of efforts to combat climate change and biodiversity loss, perpetuating environmental deterioration and vulnerability.

"Educating our youth on climate change is not just a challenge but a necessity" (Project WET, n.d.). Traditional educational methods frequently fail to engage the public effectively. Textbooks often become rapidly obsolete, lectures may be unengaging, and content might feel disconnected from people's daily experiences. This contributes to widespread gaps in public understanding of environmental challenges and potential solutions. Environmental knowledge is essential for fostering enlightened attitudes, values, and practical skills, facilitating informed decisions and sustainable actions (Rieckmann, 2018). Such education must begin early, with targeted awareness programs addressing both general and specialized groups to cultivate the skills and knowledge necessary to sustainably address environmental challenges (Project WET, n.d.).

Today's digital landscape complicates knowledge dissemination further through rampant misinformation and disinformation, particularly concerning environmental issues. "Greenwashing" practices—where companies falsely portray themselves as environmentally friendly—add confusion, challenging public trust in credible sources (Treen et al., 2020). Recent advancements in Artificial Intelligence (AI), particularly large language models (LLMs), offer promising solutions to address these informational gaps. AI chatbots, capable of processing vast datasets and generating accurate, accessible information, can significantly enhance educational effectiveness (Minaee et al., 2024). These technologies have the potential to bridge existing knowledge gaps by providing immediate, reliable environmental information to students, educators, and the general public. Furthermore, AI models can enable experiential learning through interactive simulations of climate scenarios, real-time policy analysis, and complex environmental decision-making tasks, thus making learning more engaging and practically relevant.

Various studies have highlighted AI's capability to process and teach complex environmental issues through interactive and user-friendly platforms (Rukhiran et al., 2022; Cao & Jian, 2024). AI models have been implemented to deliver tailored educational content and simulate conversation, which could be beneficial in explaining important topics like climate change and biodiversity (TeachFlow.AI, 2022). Studies focusing on AI chatbots like GPT-3.5, GPT-4, and others show that the AI models can handle a range of environmental topics with varying degrees of effectiveness. While some AI models excel in areas such as climate science, they may lack depth in others, like the socio-economic implications of environmental policies (Chi, 2024; Urzedo et al., 2024). The integration of AI models in educational systems has been debated for its potential to enhance learning outcomes and engage the public on environmental issues (AiwaBlog, n.d.; Rukhiran et al., 2022). However, the reliability of AI models in education remains a concern. The ethical application of using AI models in education, for instance, especially concerning privacy, security, and accuracy of the information, are crucial considerations (Cao & Jian, 2024).

This study explores the differences in environmental knowledge between humans and AI, providing insights into the capabilities of AI models to understand complex environmental concepts and their implications for education and policy. AI-driven methods have demonstrated potential in enhancing educational processes and generating content, which could significantly transform environmental education by making complex information more accessible and engaging. Moreover, AI can potentially facilitate a deeper understanding of environmental systems, optimize management of blue and green infrastructure, and offer advanced tools for effective environmental decision-making.

To address the existing gap in empirical research, this study systematically evaluates and compares the effectiveness of prominent AI models GPT-3.5, GPT-4, GPT-4o, Gemini (Gemini 1.0), Claude (Claude 3.5 Sonnet), and Llama 2 (Llama 2 70B) using the validated Environmental Knowledge Test (EKT-19). The investigation is guided by clearly defined research questions designed to provide actionable insights:

    RQ1: What are the comparative strengths and limitations of these models in environmental education?
    RQ2: Can AI effectively engage and educate individuals on environmental issues compared to traditional methods?
    RQ3: How do different AI models contribute to environmental education?

Through assessing these AI capabilities, the study aims to determine how these technologies can enhance understanding, support informed decision-making, and encourage sustainable practices.

The paper is structured as follows: Section 2 provides the background about the large language models employed in the paper. Section 3 describes the methodology used in the study. In Section 4, we present the results of the comparative evaluation. In Section 5, we discuss the study's findings. In Section 6, we discuss the limitations of the AI models in environmental education. Section 7 concludes the paper.

## 2. AI Models Examination

AI models excel at delivering customized quizzes and multiple-choice questions (if appropriately prompted), offering immediate feedback and personalized learning paths based on user responses. LLMs such as GPT-4 and GPT-4o boast access to immense environmental science datasets, allowing them to formulate complex, scenario-based assessments that test a student's understanding of cause-and-effect relationships in environmental issues. Additionally, AI models like LLama can be trained on specific environmental topics, enabling in-depth assessments tailored to regional challenges or environmental professions.

They can also readily generate summaries of scientific findings, present current environmental statistics, and answer factual questions with high accuracy (for instance, [https://climategpt.ai/]). However, AI models currently struggle with open-ended, creative questions that require critical thinking and real-world application of knowledge, as identified by Hultberg et al. (2024) and Calonge et al. (2023). AI models also struggle with fostering critical thinking and debate which are crucial aspects of environmental education. Additionally, the sheer volume of information available online necessitates robust fact-checking mechanisms within AI models, which are not always in place. Bias in training data can lead to the proliferation of misinformation, inaccurate responses, and hallucinations, whereby the chatbot may "invent" details to fill in gaps a person left out in the prompt. For instance, Urzedo et al. (2024) noted that "the chatbot overlooked Indigenous and community-led restoration organisations" (p.3) and "reproduced a significant technical limitation in restoration efforts by neglecting the significance of non-forest ecosystems and non-tree plant species" (p.5).

Human assessors can analyze a student's thought process, probe, question, critique, identify misconceptions, and provide nuanced constructive feedback and additional background beyond pre-programmed responses. Moreover, assessing a student's ability to analyze data (beyond description), synthesize information from various sources, and form reasoned, analytical, and structured arguments – all crucial environmental skills – remains a challenge for AI. Chatbots, assistants, and research platforms like Litmaps [https://www.litmaps.com/] or SciSpace [https://scispace.com/] excel at providing accessible and engaging environmental articles, answers, or graphs but cannot, at the moment, sustain an argument with a human.

The use of chatbots for climate change education is exemplified by a United Nations initiative that employs a Facebook Messenger bot to promote climate change understanding and urge personal action. This initiative underscores the potential of chatbots to engage a wide audience and stimulate proactive environmental behavior (UN News, 2018).

AI models can generate educational content in various formats, from summaries of scientific papers to interactive simulations. Interactive, gamified experiences offered by AI models like Llama 2 can make learning enjoyable and promote knowledge retention. Additionally, the combination of various chatbots and assistants can cater to diverse learning preferences, delivering information through text, audio, images, or video, increasing accessibility for a broader audience. However, AI models currently lack the human touch and the ability to adapt, rephrase, and refine explanations based on a student's level of understanding. The efficacy of AI models is also contingent on robust internet and WiFi infrastructure. In regions with limited access to technology, the reach and impact of AI-driven environmental education and advocacy may be significantly constrained, exacerbating the digital divide.

A skilled and discipline-expert human educator can tailor their communication style, provide and elaborate on concrete examples (lab, industry, etc.), and address individual needs. While AI models can provide information, human experts can engage in more nuanced public outreach and advocacy efforts. They can participate in forums, debates, and community initiatives, leveraging the initial engagement facilitated by AI models to drive deeper understanding and actions. Discussions with human educators enable students to challenge assumptions and biases, reflect, explore different perspectives, and develop well-rounded environmental knowledge.

In summary, AI models offer a valuable tool for assessing and enhancing environmental knowledge. Their ability to deliver accessible, interactive learning experiences and provide readily available information positions them as powerful educational aids. However, limitations in critical thinking assessment, fostering debate, and mitigating bias necessitate the continued role of human educators. The most effective approach likely lies in a collaborative model, where AI models personalize learning experiences and deliver factual information, while human educators guide critical thinking, address misconceptions, and facilitate meaningful discussions.

## 3. Methodology

In this study, the Environmental Knowledge Test (EKT-19) was used to assess general environmental knowledge (Player et al., 2023) in both the university student participants and the AI model responses. We continued to use EKT-19 as the main assessment tool because this standardized test is psychometrically validated and comprehensive, covering a broad range of environmental topics and providing a reliable benchmark for evaluating knowledge levels in our AI models and human participants. The EKT-19 was supplemented with additional custom-designed short-answer questions targeting specific environmental knowledge areas. These additional questions were included to mitigate potential training bias in the AI models and ensure assessment of knowledge beyond what is covered in EKT-19. This approach aligns with the study's broader goal of not only evaluating knowledge but also enhancing it; here, 'enhance' refers to helping learners better understand environmental concepts, gain easier access to relevant information, and apply that knowledge more effectively through the support of AI models.

### 3.1 Sampling Method
We used a convenience sampling strategy to recruit university students for the study. This non-probabilistic sampling technique was chosen to include all volunteers expressing interest during a specified announcement period, without a predefined sample size limit. A campus-wide announcement was utilized to recruit volunteers. In the end, we collected data from 46 students.

### 3.2 Participants
Students over the age of 18, from various disciplines, participated in the study. The sample included 46 participants, with 93% identifying as female. The age range was 19–22 years, with a mean of 20.5 years (SD = 0.89). Most participants were third-year students (67%), followed by fourth-year (28%) and second-year students (4%). The academic disciplines represented included Environmental Science and Sustainability (52%), Computer Science, Information Technology, Media, Psychology, Political Sciences, Social Innovation, Business, Marketing, and International Relations (48%). Recruitment was facilitated through a detailed announcement that outlined the study's purpose and requirements.This announcement also stressed the voluntary nature of participation, including the option for participants to withdraw at any time without penalty. Informed consent was sought. No financial incentives were offered.

### 3.3 Instruments
The primary instrument used in the study was the Environmental Knowledge Test-EKT-19 (Player et al., 2023), a publicly available standardized test consisting of multiple-choice questions across several environmental domains: Ecology (5 items), Climate (5 items), Resources (3 items), Consumption Behavior (8 items), Society & Politics (3 items), Economy (2 items), and Environmental Contamination (4 items).

The EKT-19 is structured around three specific types of knowledge: System, Action-related, and Effectiveness knowledge, to assess environmental understanding comprehensively. These categories help differentiate the types of environmental understanding necessary for effective environmental education and behavior. First, system knowledge deals with how systems operate, including ecological processes, climate mechanisms, and resources cycles. Second, action-Related knowledge deals with the practical steps individuals, communities and governments can take to mitigate environmental problems; examples include, but not limited to: recycling, conservation efforts, and sustainable practices. The third and last point is about effectiveness knowledge, which evaluates the impact of the above-mentioned actions with a focus on their efficiency in reducing carbon footprints, conserving energy, and promoting sustainability. To address potential training biases of AI models, the EKT-19 was supplemented with three targeted short-answer questions focusing on specific areas of environmental knowledge.

### 3.4 Procedure
Students completed the EKT-19 individually in a controlled classroom setting under standardized conditions, including a set time limit. This setup ensured consistency in the administration of the test across all participants

**3.5 Ethical Clearance:** Ethical clearance for this study was obtained from the Research Ethics Committee at X University [XU24_044_F], United Arab Emirates. Prior to the test, participants were briefed on the assessment's nature and provided with resources to manage potential stress or emotional discomfort.

### 3.6 Evaluations procedure
All AI models were given the exact same test questions as the students, using identical prompts to ensure comparability. The multiple-choice questions were asked in a single prompt, while each short-answer question was prompted individually. The EKT-19 responses were graded based on the answer sheet. The responses to short-answer questions were graded independently by a field expert based on a standard rubric. Using the same assessment for both students and AI allowed a direct comparison of performance. This parallel evaluation against student performance is crucial to determine the relevance and value of AI models in an educational context, illustrating whether an AI can match or exceed human knowledge levels—a necessary consideration if such models are to be used to enhance environmental learning.

### 3.6 Data Analysis

The EKT-19 data was analyzed using descriptive statistics to calculate overall scores and identify areas of strength and weakness in environmental knowledge for the AI models and student groups. Independent t-tests were conducted to compare the AI models' scores on both the multiple-choice questions (MCQs) of the EKT-19 and short-answer questions (SAQs). Furthermore, independent t-tests were used to compare students versus AI models and between different student groups (Environmental students vs. other majors). To determine whether there was a statistically significant difference in the overall proportion of correct answers between the AI models, a chi-squared test was performed.

The details of six AI models responses to the EKT-19 test questions, along with the correct answers have been made publicly available on GitHub [https://github.com/lyndaSm/ENV-AI].

## 4. Results
### 4.1 AI models performance

The performance of AI models (GPT-3.5, GPT-4, GPT-4o, Gemini, Claude, and Llama 2) on the EKT-19 was analyzed across the following domains: ecology, climate, resources, consumption behavior, society & politics, economy, and environmental contamination. The total score on the test is 30 points. Table 1 illustrates the scores of each AI model across the different EKT-19 environmental domains. Claude achieved the highest score (30), while Llama 2 had the lowest score (22). The other AI models, GPT-4, GPT-4o, and Gemini each scored 28, and GPT-3.5 scored 24.

Table 1. Total Scores by Domain for Each AI Model

|  | GPT-3.5 | GPT-4 | GPT-4o | Gemini | Claude | Llama 2 |
|---|---|---|---|---|---|---|
| Ecology (5 items) | 3 | 4 | 4 | 5 | 5 | 2 |
| Climate (5 items) | 5 | 5 | 5 | 5 | 5 | 3 |
| Resources (3 items) | 2 | 3 | 3 | 3 | 3 | 2 |
| Consumption Behaviour (8 items) | 6 | 7 | 7 | 7 | 8 | 8 |
| Society & Politics (3 items) | 2 | 3 | 3 | 3 | 3 | 2 |
| Economy (2 items) | 2 | 2 | 2 | 2 | 2 | 2 |
| Environmental Contamination (4 items) | 4 | 4 | 4 | 3 | 4 | 3 |
| Total Score (out of 30) | 24 | 28 | 28 | 28 | 30 | 22 |

Claude performed perfectly in all domains, while Llama 2 showed variability, with high performance in Consumption Behavior and Economy but lower scores in Ecology and Climate (Figure 1).

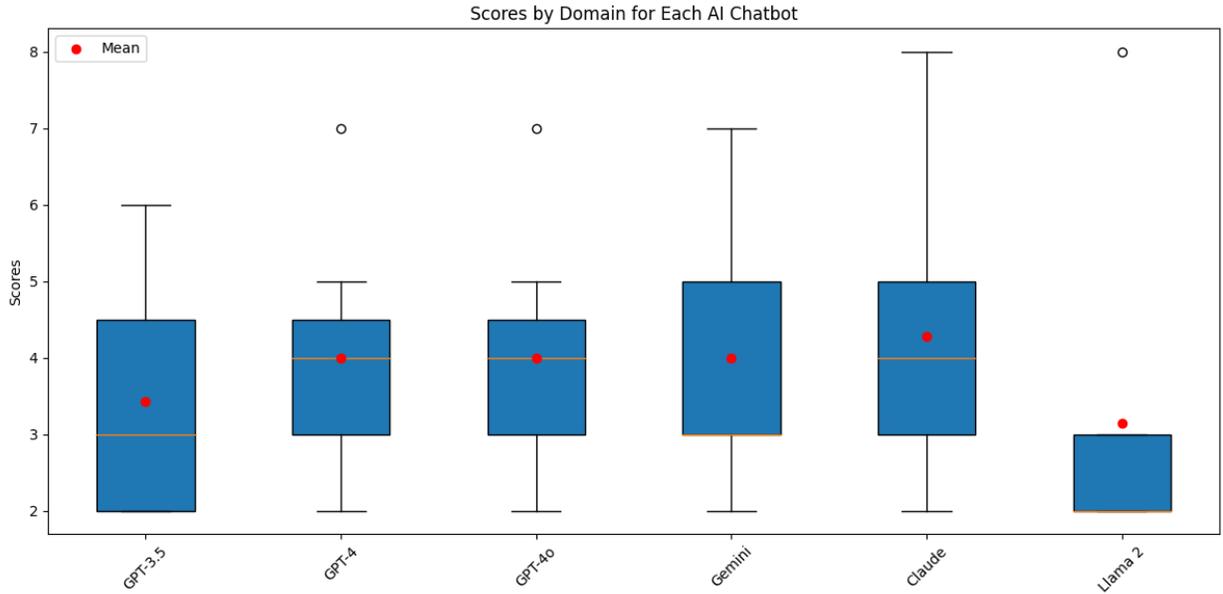

Figure 1. Boxplot of AI Models' Scores by Domain.

Table 2 indicates that Claude again performed exceptionally, scoring the highest in all three specific types of knowledge: System, Action, and Effectiveness, while Llama 2 lags behind all other AI modes, particularly in the System and Action knowledge.

Table 2. System, Action, and Effectiveness Scores for Each AI Model

|  | GPT-3.5 | GPT-4 | GPT-4o | Gemini | Claude | Llama 2 |
| --- | --- | --- | --- | --- | --- | --- |
| System (16 items) | 14 | 15 | 15 | 16 | 16 | 11 |
| Action (6 items) | 5 | 6 | 6 | 5 | 6 | 4 |
| Effectiveness (8 items) | 5 | 7 | 7 | 7 | 8 | 7 |
| Total Score | 24 | 28 | 28 | 28 | 30 | 22 |

Table 3 provides the scores for short answer questions (SAQs), with Llama 2 scoring the highest average (5) across all three questions. GPT-3.5 had the lowest average score (4.3).

Table 3. Short-Answer Questions Scores for Each AI Model

|  | GPT-3.5 | GPT-4 | GPT-4o | Gemini | Claude | Llama 2 |
| --- | --- | --- | --- | --- | --- | --- |
| SAQ 1 | 5 | 5 | 5 | 5 | 5 | 5 |
| SAQ 2 | 4 | 4 | 4 | 4 | 5 | 5 |
| SAQ 3 | 4 | 5 | 5 | 4 | 4 | 5 |
| SAQs Average | 4.3 | 4.7 | 4.7 | 4.3 | 4.7 | 5 |

The word counts from Figure 2 show considerable variability in verbosity among the different AI models. For instance, GPT-3.5 and Gemini consistently provided the shortest responses across all three SAQs (a mean of 44 words), while Llama 2 generally produced the longest responses (mean of 74 words).

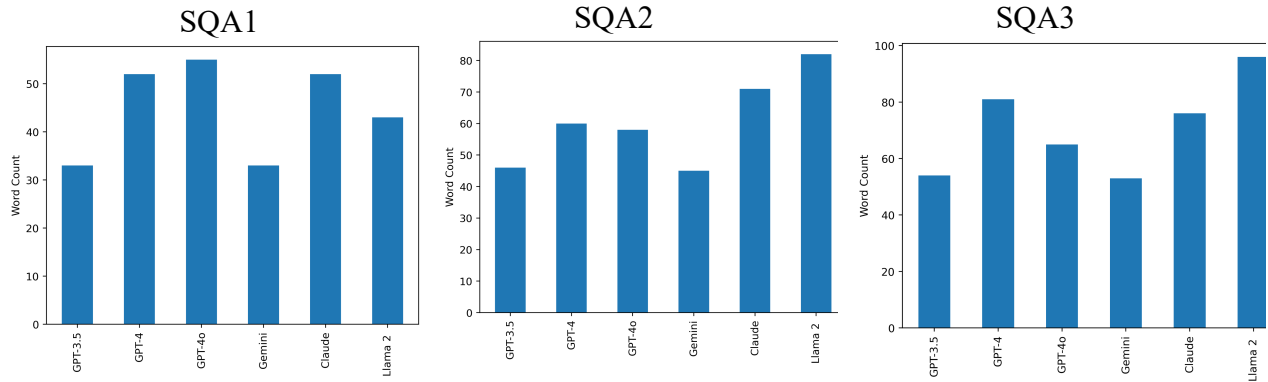

Figure 2. Word Count for Short Answer Questions

The results of the t-test are presented in Table 4. In the t-test, we conducted a pair-wise comparison of AI models. As can be seen from Table 4, there are significant differences between GPT-3.5 and Claude (p = 0.009), GPT-4 and Llama 2 (p = 0.04), GPT-4o and Llama 2 (p = 0.04), and Claude and Llama 2 (p = 0.002). The remaining pairs of AI models showed no significant differences in performance at the p-value of 0.05.

Table 4. T-Test Results Comparing Performance Differences Among AI Models

| AI Model 1 | AI Model 2 | t-statistic | p-value |
| --- | --- | --- | --- |
| GPT-3.5 | GPT-4 | -1.52 | 0.133 |
| GPT-3.5 | GPT-4o | -1.52 | 0.133 |
| GPT-3.5 | Gemini | -1.52 | 0.133 |
| GPT-3.5 | Claude | -2.69 | 0.009* |
| GPT-3.5 | Llama 2 | 0.60 | 0.549 |
| GPT-4 | GPT-4o | 0 | 1 |
| GPT-4 | Gemini | 0 | 1 |
| GPT-4 | Claude | -1.44 | 0.155 |
| GPT-4 | Llama 2 | 2.12 | 0.04* |
| GPT-4o | Gemini | 0 | 1 |
| GPT-4o | Claude | -1.44 | 0.155 |
| GPT-4o | Llama 2 | 2.12 | 0.04* |
| Gemini | Claude | -1.44 | 0.155 |
| Gemini | Llama 2 | 2.12 | 0.04* |
| Claude | Llama 2 | 3.25 | 0.002* |

A chi-squared test was performed to determine whether there is a statistically significant difference in the overall proportion of correct answers between the AI models. In other words, we tested the null hypothesis: $p_1=p_2=\ldots=p_6$, where $p_i$ is the proportion of correct answers by the ith AI model. The chi-squared statistic score ($\chi^2$) was 15.30, with the p-value of 0.0092. The p-value is smaller than the standard significance level of 0.05 (or even 0.01), indicating a statistically significant difference in model performance in terms of correct and incorrect counts.

To further explore the differences in performance, we carried out a pair-wise comparison of the models using the Marascuillo procedure at alpha 0.05. The results in Table 5 show that there is no statistically significant difference between individual models. However, the result of the last pair is very close, suggesting a real difference between the Claude and Llama models. The limited availability of data constraints the ability to obtain more robust results. Nevertheless, the preliminary results do show that the models have significantly different performances.

Table 5. The Marascuillo procedure at alpha 0.05

| AI Model 1 | AI Model 2 | $|p_i-p_j|$ | $r_{ij}$ |
| --- | --- | --- | --- |
| GPT-3.5 | GPT-4 | 0.13 | 0.2882 |
| GPT-3.5 | GPT-4o | 0.13 | 0.2882 |
| GPT-3.5 | Gemini | 0.13 | 0.2882 |
| GPT-3.5 | Claude | 0.20 | 0.2411 |
| GPT-3.5 | Llama 2 | 0.07 | 0.3376 |
| GPT-4 | GPT-4o | 0 | 0.1524 |
| GPT-4 | Gemini | 0 | 0.1524 |
| GPT-4 | Claude | 0.07 | 0.0961 |
| GPT-4 | Llama 2 | 0.20 | 0.3376 |
| GPT-4o | Gemini | 0 | 0.1524 |
| GPT-4o | Claude | 0.07 | 0.0961 |
| GPT-4o | Llama 2 | 0.20 | 0.3376 |
| Gemini | Claude | 0.07 | 0.0961 |
| Gemini | Llama 2 | 0.20 | 0.3376 |
| Claude | Llama 2 | 0.27 | 0.2771 |

**4.2 Students performance vs. AI models**

The study included 46 students with mostly female participants (93%). The age range was 19–22 years, with mean 20.5 and standard deviation of 0.89. The majority of participants were in their third year of study (67%), followed by fourth-year (28%) and second-year students (4%). The majority (65%) came from the College of Natural and Health Sciences, with smaller percentages from other colleges, such as the College of Technological Innovation and the College of Interdisciplinary Studies (11% each). In terms of field of major, 52% of students were enrolled in the Environmental Science and Sustainability program, while 48% studied Computer Science, Information Technology, Media, Psychology, Political Sciences, Social Innovation, Business, Marketing, or International Relations.

The students' mean score on the EKT-19 was 13.20 (SD=3.98), while the AI models' mean score 26.67 (SD=3.01) was significantly higher (Table 6). The median score was 12 for students and 28 for AI models, indicating a substantial difference in central tendencies.

AI models outperformed students across all environmental domains (Figure 3). The most significant differences were observed in the domains of resources (students: 0.87 vs. AI models: 2.67) and consumption behavior (students: 3.61 vs. AI models: 7.17).

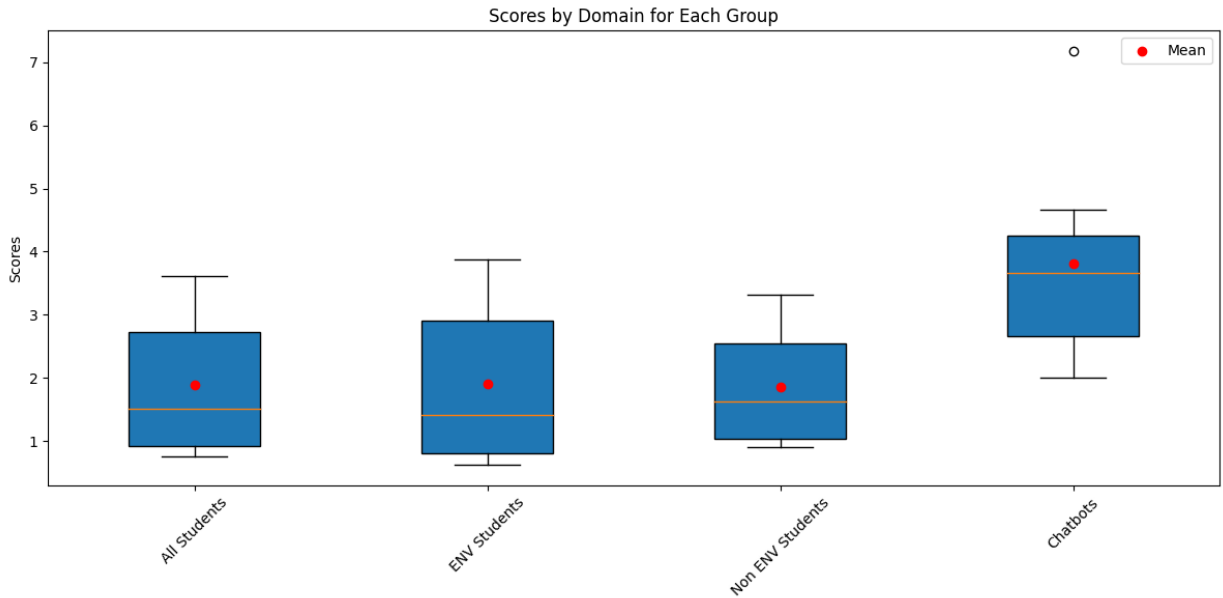

Figure 3. Boxplot of All Students, ENV, Non ENV students, and AI Models' Scores by Domain.

For the EKT-19 scores by each domain of knowledge, the results showed that AI models again surpassed student performance in the three domains: System, Action, and Effectiveness. The short answer question (SAQ) performance showed similar trends. Students had an overall average of 2.22, while AI models had an overall average of 4.61. This higher performance in SAQs indicates AI models' better ability to generate comprehensive and accurate responses to open-ended questions.

Table 6. EKT-19 Scores per Domain Comparison

|  | All Students | ENV Students | Non-ENV Students | AI Models |
|---|---|---|---|---|
| Ecology (5 items) | 2.46 | 2.67 | 2.23 | 3.83 |
| Climate (5 items) | 3.00 | 3.13 | 2.86 | 4.67 |
| Resources (3 items) | 0.87 | 0.79 | 0.95 | 2.67 |
| Consumption Behaviour (8 items) | 3.61 | 3.88 | 3.32 | 7.17 |
| Society & Politics (3 items) | 0.98 | 0.83 | 1.14 | 2.67 |
| Economy (2 items) | 0.76 | 0.63 | 0.91 | 2.00 |
| Environmental Contamination (4 items) | 1.51 | 1.42 | 1.62 | 3.67 |
| EKT-19 test overall mean | 13.20 | 13.33 | 13.05 | 26.67 |

Comparing environmental science (ENV) students to non-ENV students revealed slight performance differences. ENV students had a mean MCQ score of 13.33 (SD = 3.55), slightly higher than non-ENV students' mean of 13.05 (SD = 4.48). In domain-specific scores, ENV students performed better in Ecology (2.67 vs. 2.23) and Climate (3.13 vs. 2.86), while non-ENV students scored slightly higher in Resources (0.95 vs. 0.79). Moreover, System, action, and effectiveness scores were comparable between the groups, with minor differences favoring ENV students in some areas (Table 7).

Table 7. System, Action, and Effectiveness Scores Comparison

| | All Students | ENV Students | Non-ENV Students | AI Models |
|---|---|---|---|---|
| System (16 items) | 7.67 | 7.67 | 7.68 | 14.50 |
| Action (6 items) | 3.20 | 3.13 | 3.27 | 6.33 |
| Effectiveness (8 items) | 2.33 | 2.54 | 2.09 | 6.83 |

The t-tests conducted to compare the scores of different groups across various environmental domains showed that AI models significantly outperform across student groups: all students (t = -2.46, p-value = 0.03), ENV students (t = -2.32, p-value = 0.04), and non-ENV students (t = -2.61, p-value = 0.02). This shows that AI models can be valuable tools for improving environmental education. There is no significant difference in performance between ENV and non-ENV students (t = 0.07, p-value = 0.94). A probable explanation for the high p-value is that the sample size was insufficient to detect a meaningful difference, even if one existed.

5. Discussion

The results of the study reveal a considerable potential of AI models to enhance environmental knowledge among students. AI models, including GPT-3.5, GPT-4, GPT-4o, Gemini, Claude, and Llama 2, outperformed all students across various domains of the EKT-19, including ecology, climate, resources, consumption behavior, society & politics, economy, and environmental contamination. The results indicate that AI models can be useful in conveying environmental concepts and giving immediate and accurate information, which allows them to supplement traditional education. However, the findings also underscored the challenges facing some AI models in handling open-ended questions and the need for human oversight to maintain both the accuracy and contextual relevance of the information they deliver. AI models have access to large volumes of data and can be trained on specific types of environmental data, such as scientific research, reports, and real-time monitoring information. This will allow AI modes to potentially recall specific details and statistics more accurately than a human, particularly if there were no opportunities for retrieval practice and spacing for students in class (Hultberg et al., 2018) before taking the EKT-19.

The results showed that AI models had a significantly higher mean score (26.67 or 88.89%) compared to students (13.20 or 43.99%) on the EKT-19. This performance gap was evident across all specific knowledge domains, indicating that AI models can comprehensively cover a wide range of environmental topics.

*RQ1: What are the comparative strengths and limitations of these models in environmental education?*

Several models have demonstrated impressive performance on the EKT-19, indicating their potential strength in environmental education. Notably, GPT-4, GPT-4o, and Gemini achieved an overall score of 93%, while Claude achieved a remarkable 100% on the EKT-19. This performance suggests that the AI models have attained a near-human-level understanding of environmental concepts, opening up numerous possibilities for the future applications of language models in environmental education. Just a few years ago, it was inconceivable that a machine learning model could comprehend natural language text and respond effectively. Therefore, it is expected that further advancements in machine learning technology will continue to improve language model performance in environmental education.

The AI models performed particularly well in the categories of Climate, Resources, Society & Politics, Economy, and Environmental Contamination, with several models obtaining perfect scores. This success can be attributed to the availability of extensive public data related to these topics. Generally, more training data leads to better model performance. The consistent high performance across multiple models supports this hypothesis about the availability of public data. To enhance performance in categories with lower scores, such as Ecology and Consumption Behavior, researchers must provide more training data for the machine learning algorithms.

Climate was the best-performing category, with all but one model achieving perfect scores. This is likely due to the abundance of text and information available on the internet related to climate, a well-researched subject with a large corpus of open-access research that can be used for training language models. Economy is also a strong subject area, which is surprising given the availability of information on this fundamental topic in science. The significant amount of information on the economy in the context of the environment facilitates the training of language models (TeachFlow.AI, 2022).

The Claude model stands out with its exceptional performance, using constitutional AI, an approach developed by Anthropic for training AI systems to be harmless and helpful without extensive human feedback. The results indicate that Claude's architecture is well-suited for environmental education, suggesting that constitutional AI may be more effective than other approaches in the context of environmental theory.

While all tested models performed relatively well on the EKT-19, GPT-3.5 and Llama 2 underperformed compared to the rest of the cohort, with scores of 80% and 73.3%, respectively.

These results align with multiple other studies (Hultberg et al., 2024; Calonge et al., 2023) showing that Llama 2 lags behind other state-of-the-art models. As an open-source model, Llama 2 lacks the same backing as its commercial peers, contributing to its underwhelming scores. However, unlike commercial closed-source models, Llama 2 can be specifically fine-tuned to a particular topic to improve its knowledge and understanding.

Ecology was the weakest area of knowledge for most models. Except for Claude and Gemini, all models received suboptimal scores on this topic. The results in Ecology are somewhat surprising compared to Climate, as the two subjects are closely related. Given the similarity in content and availability of data for both topics, the difference in performance may be due to the internal architecture of the GPT and Llama models (Minaee et al., 2024).

*RQ2: Can AI effectively engage and educate individuals on environmental issues compared to traditional methods?*

Imagine having access to an indefatigable virtual environmental expert 24/7. AI models can provide consistent answers every time, regardless of mood or exhaustion. That is the promise of commercially-(and nearly freely) available AI models. They can provide instant and well-researched answers to questions about climate change, pollution, conservation, and more. They can also evaluate (if prompted and by using prompt engineering techniques such as Few-Shot-Chain-of-Thought), explain, and simplify complex concepts, making environmental knowledge accessible to everyone.

This study's findings support these claims, demonstrating empirically that AI models significantly outperform traditional student learning as evidenced by superior performance on the EKT-19 assessments. Specifically, AI models achieved a mean score of 26.67 (88.89%), significantly higher than the student participants' mean score of 13.20 (43.99%), demonstrating their clear advantage in conveying comprehensive and accurate environmental information. Additionally, AI models consistently outperformed students across all environmental domains evaluated, including Ecology, Climate, Resources, Consumption Behavior, Society & Politics, Economy, and Environmental Contamination. This consistent performance underscores AI's broad capability to cover diverse environmental issues effectively.

Lectures and texts are common examples of passive learning methods used in traditional schooling. In contrast, AI models improve engagement through quizzes, games, and simulations. However, the real transformative potential lies in their ability to create interactive, systems-based simulations that allow learners to explore real-world environmental challenges, such as ecosystem management or climate adaptation strategies. Such applications could foster systems thinking and critical pedagogy, enabling students to understand interconnections and consequences within complex environmental systems (Rukhiran et al., 2022; Cao & Jian, 2024).

AI-driven platforms can further transform environmental education by fostering experiential learning. These platforms can create virtual ecosystems where students actively engage with complex scenarios, such as implementing renewable energy policies or designing water conservation strategies. By simulating real-world challenges, students can explore the outcomes of their decisions in a risk-free environment, fostering deeper engagement and understanding. For example, an AI-powered simulation could model the effects of deforestation on biodiversity and carbon emissions, allowing learners to test reforestation policies and analyze their effectiveness.

Beyond providing engaging and immersive experiences, AI can personalize environmental education by identifying individual learners' weaknesses and adapting content accordingly. One of its most promising applications is the ability to provide tailored feedback, customized learning pathways, and targeted resources, helping students address gaps in understanding at their own pace. Additionally, AI-generated quizzes and gamified learning tools can boost retention and motivation by dynamically adjusting difficulty levels based on learner progress, creating a more responsive and adaptive educational experience.

Several studies documented the advantages of AI in educational settings. AI-based teaching systems have demonstrated their capacity to improve environmental knowledge and attitudes among students, showing that AI can engage learners more effectively than traditional methods (Huang, 2018). AI's ability to democratize access to education is particularly crucial for reaching underrepresented groups who may not have access to traditional educational resources, provided that they have access to Internet/wifi. AI tools can provide personalized and interactive simulations that adjust to individual learning paces and provide real-time feedback, thereby improving the learning experience.

Additionally, AI has the ability to handle and analyze massive volumes of data, allowing the generation of well-designed educational content on a wide range of environmental themes. AI models can keep up with the latest scientific research, ensuring that educational materials are current, relevant, and correct (Islam et al., 2024). However, to fully leverage the educational benefits of AI models, it is crucial to ensure that these AI tools are available/accessible to all students, regardless of their socio-economic status. AI models have the potential to improve environmental education by making it more engaging, fun, accessible, and up-to-date. Educators therefore need sufficient continuing professional development to effectively integrate AI tools into their teaching approaches (Huang, 2018).

Empirical results from this study further support these assertions, as AI models significantly surpassed traditional student performance on standardized environmental knowledge assessments (EKT-19), indicating greater effectiveness in delivering precise and consistent environmental information. Specifically, AI models achieved a mean score of 26.67 (88.89%), significantly higher than the student participants' mean score of 13.20 (43.99%), demonstrating their clear advantage in conveying comprehensive and accurate environmental information. Additionally, AI models consistently outperformed students across all environmental domains evaluated, including Ecology, Climate, Resources, Consumption Behavior, Society & Politics, Economy, and Environmental Contamination. This consistent performance underscores AI's broad capability to cover diverse environmental issues effectively.

Nonetheless, the integration of human expertise remains essential to address nuanced questions, promote critical thinking, and facilitate meaningful discussions. Future research should explicitly evaluate user engagement using qualitative methods such as interviews and surveys, as well as longitudinal studies to assess long-term knowledge retention and behavioral change resulting from interactions with AI models compared to traditional instructional methods. Additionally, emphasizing the complementary role of AI alongside human educators could further enhance educational outcomes by combining personalized, interactive content delivery with human-facilitated critical thinking and reflection.

*RQ3: How do different AI models contribute to environmental education?*

To address RQ3 clearly and effectively, this study assessed the performance of various prominent AI models, including GPT-3.5, GPT-4, GPT-4o, Gemini, Claude, and Llama 2, using the standardized Environmental Knowledge Test (EKT-19). The results indicate that these AI models offer distinct but complementary contributions to environmental education. For instance, Claude demonstrated exceptional capability, scoring a perfect 100% on the EKT-19, reflecting its superior performance and reliability in delivering accurate and comprehensive environmental information. In contrast, Llama 2, despite its lower overall score (73.3%), excelled in generating detailed responses in open-ended questions, indicating its potential strength in handling complex and nuanced environmental discussions.

Moreover, GPT-4, GPT-4o, and Gemini consistently performed strongly across diverse environmental domains, showing their capability to cover a wide range of environmental topics effectively. The study highlights that different models have unique strengths, which can be strategically leveraged depending on educational needs and contexts. For example, models like GPT-4 and Claude, due to their accuracy and consistency, can be highly effective for content dissemination and routine educational queries, while Llama 2's capacity for detailed explanations could enhance critical thinking and engagement in more complex environmental topics.

This study underscores the importance of integrating these diverse capabilities within a cohesive educational strategy. A collaborative approach combining the strengths of AI models with human educators can optimize environmental education. Educators and students can create engaging content on environmental topics, such as articles, videos, and infographics, while AI models can distribute this content widely via various platforms, including websites and social media. AI models can also manage routine inquiries during interactive Q&A sessions, allowing educators to focus on complex, nuanced discussions.

Furthermore, AI models significantly enhance environmental education by providing personalized learning experiences. Educators can design personalized learning pathways tailored to individual knowledge levels and interests, with AI models dynamically guiding learners through these pathways by offering real-time feedback and relevant resources. Interactive simulations powered by AI models can facilitate experiential learning, making abstract environmental concepts tangible and relatable.

Additionally, AI models support educators by collecting and analyzing data on learner interactions, common misconceptions, and frequently asked questions. This valuable insight enables educators to refine teaching strategies, develop targeted interventions, and continually improve the effectiveness of environmental education programs. AI models can also assist educators and students in community-building activities, moderating discussions, and providing real-time information during webinars and forums. In awareness campaigns, educators define key messages and goals, while AI models execute these campaigns through personalized communications, reminders, and feedback collection to refine future content.

In conclusion, while AI models individually offer substantial contributions—ranging from superior accuracy in environmental knowledge delivery to nuanced engagement in complex environmental discussions—the most impactful approach involves strategically integrating these diverse strengths with human educational expertise. This integrated approach ensures a balanced, comprehensive, and highly effective environmental education that enhances both learner knowledge and critical thinking skills.

## 6. Study Limitation

This study has several limitations. First, the use of convenience sampling restricts the representativeness of the sample, potentially introducing selection bias and limiting the generalizability of findings. The small sample size of 46 university students from the United Arab Emirates further constrains statistical power, making it challenging to extend results to a broader population. Additionally, focusing exclusively on university students limits applicability to other demographic groups. Second, the cross-sectional design emphasizes immediate knowledge gains but does not address long-term learning outcomes, such as skills development, critical thinking enhancement, or real-world knowledge application. Future research should therefore employ longitudinal methodologies to comprehensively assess these important dimensions. Third, the reliance on the EKT-19, although standardized, might not fully encompass all aspects of environmental knowledge, while supplementary short-answer questions, designed to mitigate AI training biases, might introduce subjectivity in scoring. Fourth, AI model adaptability and continuous learning capabilities could benefit from additional training, ensuring up-to-date accuracy with the rapidly evolving environmental science landscape. Finally, the small sample size and potential confounding variables limit the reliability of statistical analyses, even though the applied independent t-tests and chi-squared tests are appropriate. Future studies should aim to overcome these limitations by employing more robust sampling methods, increasing sample sizes, and refining assessment instruments and procedures.

## 7. Conclusion and Future Research

In conclusion, AI models hold significant promise in bridging the gap in environmental knowledge not only between experts and the public but also between educators and students, as out of class activity on climate change, climate action and carbon footprint for instance, as indicated by Menkhoff and Gan (2024). By providing accessible, accurate (Thulke et al., 2024), and timely information, these agents/assistants can enhance environmental literacy and empower individuals to make informed decisions about sustainability. However, it is crucial to address challenges such as ensuring data accuracy as highlighted by Urzedo et al. (2024), mitigating biases, and fostering user trust. As AI technology advances, ongoing research and interdisciplinary collaboration will be essential to maximize the potential of AI models in promoting environmental awareness and action, ultimately contributing to a more informed and engaged society.

By working synergistically, in a collaborative hybrid human-chatbots partnership model (Barany et al., 2024), we can create a more robust and responsive system to address environmental issues, where AI models serve as a first point of access to information, while human experts elaborate and provide deeper insights and guide informed decision-making. This partnership leverages the strengths of both human insights and artificial intelligence, hopefully fostering a more informed and proactive approach to environmental stewardship, awareness, and consciousness.

Beyond factual knowledge transfer, future research should investigate how these tools can actively shape learning through personalized learning experiences, fostering critical thinking, and enabling collaborative problem-solving. This includes integrating AI with real-world applications, such as policy analysis tools that allow students to simulate the effects of environmental decisions, or using AI to guide students in designing and implementing sustainability projects. Furthermore, exploring AI's role in enhancing interdisciplinary critical thinking and creativity would provide valuable insights into its broader educational potential.

**Declarations:**

**Author Contributions**: Conceptualization: LS; Data Curation: LS, FK; Methodology: LS, NO; Formal Analysis: LS, FK; Writing-original draft: LS, DSC, FK, NO; Writing-review and editing: LS, DSC, FK. All authors have read and agreed to the published version of the manuscript.

**Data Availability Statement:** The datasets used/analyzed during the current study are available from the corresponding author upon reasonable request.

**Funding**
This research received no external funding.

**Conflict of interest statement**
On behalf of all authors, the corresponding author states that there is no conflict of interest.

## References

1. AiwaBlog. (n.d.). Environmental awareness: AiwaGPT's sustainability chatbot for eco-friendly tips. Medium. Retrieved from https://medium.com/@AiwaBlog/environmental-awareness-aiwagpts-sustainability-chatbot-for-eco-friendly-tips-8dd748735513
2. Barany, A., Nasiar, N., Porter, C., & Baker, R. (2024, July). ChatGPT for education research: Exploring the potential of large language models for qualitative codebook development. Paper presented at the 25th International Conference on Artificial Intelligence in Education, Recife, Brazil.

3. Calonge, D. S., Smail, L., & Kamalov, F. (2023). Enough of the chit-chat: A comparative analysis of four AI chatbots for calculus and statistics. *Journal of Applied Learning and Teaching*, *6*(2). https://doi.org/10.37074/jalt.2023.6.2.22
4. Cao, E., & Jian, Y. (2024). The role of integrating AI and VR in fostering environmental awareness and enhancing activism among college students. Science of The Total Environment, 908, 168200. https://doi.org/10.1016/j.scitotenv.2023.168200
5. Chetan-Welsh, H., & Hendry, L. (n.d.). How are climate change and biodiversity loss linked? Natural History Museum. Retrieved from https://www.nhm.ac.uk/discover/how-are-climate-change-and-biodiversity-loss-linked.html
6. Chi, N. T. K. (2024). The Effect of AI Chatbots on Pro-environment Attitude and Willingness to Pay for Environment Protection. Sage Open, 14(1). https://doi.org/10.1177/21582440231226001
7. Hultberg, P. T., Santandreu Calonge, D., Kamalov, F., & Smail, L. (2024). Comparing and assessing four AI chatbots' competence in economics. Plos one, 19(5), e0297804. https://doi.org/10.1371/journal.pone.0297804
8. Hultberg, P., Calonge, D. S., & Lee, A. E. S. (2018). Promoting long-lasting learning through instructional design. Journal of the Scholarship of Teaching and Learning, 18(3). https://doi.org/10.14434/josotl.v18i3.23179
9. Huang, S.-P. (2018). Effects of Using Artificial Intelligence Teaching System for Environmental Education on Environmental Knowledge and Attitude. Eurasia Journal of Mathematics, Science and Technology Education, 14(7), 3277-3284. https://doi.org/10.29333/ejmste/91248
10. Masson-Delmotte, V., Zhai, P., Pörtner, H.-O., Roberts, D., Skea, J., Shukla, P. R., Pirani, A., Moufouma-Okia, W., Péan, C., Pidcock, R., Connors, S., Matthews, J. B. R., Chen, Y., Zhou, X., Gomis, M. I., Lonnoy, E., Maycock, T., Tignor, M., & Waterfield, T. (Eds.). (2018). Global warming of 1.5°C. An IPCC special report on the impacts of global warming of 1.5°C above pre-industrial levels and related global greenhouse gas emission pathways, in the context of strengthening the global response to the threat of climate change, sustainable development, and efforts to eradicate poverty. Cambridge University Press. https://doi.org/10.1017/9781009157940
11. Kamalov, F., Santandreu Calonge, D., & Gurrib, I. (2023). New era of artificial intelligence in education: Towards a sustainable multifaceted revolution. Sustainability, 15(16), 12451. https://doi.org/10.3390/su151612451
12. Iberdrola. (n.d.). The importance of climate change education. Retrieved May 18, 2024, from https://www.iberdrola.com/social-commitment/climate-change-education
13. Islam, Z., Ahmed, A., Alfify, M. H., & Riyaz, N. (2024). The impact of artificial intelligence on environment and sustainable development in India. Educational Administration: Theory and Practice, 30(5), 1850-1856. https://doi.org/10.53555/kuey.v30i5.3196

14. Minaee, S., Mikolov, T., Nikzad, N., Chenaghlu, M., Socher, R., Amatriain, X., & Gao, J. (2024). Large language models: A survey. arXiv. https://arxiv.org/html/2402.06196v2
15. Menkhoff, T., & Gan, B. (2023). Engaging students through conversational chatbots and digital content: A climate action perspective. https://doi.org/10.54941/ahfe1002960
16. Nature, biodiversity and health: an overview of interconnections. Copenhagen: WHO Regional Office for Europe; 2021. Licence: CC BY-NC-SA 3.0 IGO.
17. Player, L., Hanel, P. H. P., Whitmarsh, L., & Shah, P. (2023). The 19-Item Environmental Knowledge Test (EKT-19): A short, psychometrically robust measure of environmental knowledge. Heliyon, 9(8), e17862. https://doi.org/10.1016/j.heliyon.2023.e17862.
10. Project WET. (n.d.). Educating our youth on climate change: Strategies and resources. Retrieved May 18, 2024, from https://www.projectwet.org/blog/educating-our-youth-climate-change-strategies-and-resources
11. Rieckmann, M. (2018). Learning to transform the world: Key competencies in education for sustainable development. In A. Leicht, J. Heiss, & W. J. Byun (Eds.), Issues and trends in education for sustainable development (pp. 39-59). Paris: UNESCO. https://unesdoc.unesco.org/ark:/48223/pf0000261802
12. Rukhiran, M., Phaokla, N., & Netinant, P. (2022). Adoption of environmental information chatbot services based on the internet of educational things in smart schools: Structural equation modeling approach. Sustainability, 14(23), 15621. https://doi.org/10.3390/su142315621
13. TeachFlow.AI. (2022, November 29). The role of AI in promoting environmental education. Retrieved from https://teachflow.ai/the-role-of-ai-in-promoting-environmental-education/
14. Thulke, D., Gao, Y., Pelser, P., Brune, R., Jalota, R., Fok, F., ... & Erasmus, D. (2024). ClimateGPT: Towards AI Synthesizing Interdisciplinary Research on Climate Change. *arXiv preprint arXiv:2401.09646*. https://doi.org/10.48550/arXiv.2401.09646
15. Treen, K. M. I., Williams, H. T. P., & O'Neill, S. J. (2020). Online misinformation about climate change. WIREs Climate Change, 11(5), e665. https://doi.org/10.1002/wcc.665
16. United Nations. (n.d.). Biodiversity, our strongest natural defense against climate change. Retrieved from https://www.un.org/en/climatechange/science/climate-issues/biodiversity
17. United Nations. (n.d.). Education is key to addressing climate change. Retrieved May 18, 2024, from https://www.un.org/en/climatechange/climate-solutions/education-key-addressing-climate-change
18. United Nations. (2018, December 3). UN launches Facebook Messenger-powered bot to take on climate change. UN News. https://news.un.org/en/story/2018/12/1027471
19. Urzedo, D., Sworna, Z. T., Hoskins, A. J., et al. (2024). AI chatbots contribute to global conservation injustices. Humanities & Social Sciences Communications, 11(204). https://doi.org/10.1057/s41599-024-02720-3